\documentclass[conference,a4paper]{IEEEtran}
\IEEEoverridecommandlockouts
\usepackage{url}
\usepackage{cite}
\usepackage{amsmath,amssymb,amsfonts}
\usepackage{algorithmic}
\usepackage{graphicx}
\usepackage{textcomp}
\usepackage{stfloats}
\usepackage{xcolor}
\usepackage{array}
\usepackage{needspace}
\usepackage{threeparttable}
\usepackage{tabularx}
\usepackage{booktabs}
\usepackage{ragged2e}
\usepackage{algorithm}
\usepackage{hyperref}
\def\BibTeX{{\rm B\kern-.05em{\sc i\kern-.025em b}\kern-.08em
    T\kern-.1667em\lower.7ex\hbox{E}\kern-.125emX}}

\newcommand{\etal}{\textit{et al.}}

\begin{document}

\title{Retinal OCTA Phenotyping with LLM Reporting for Alzheimer’s Disease}

\author{
\IEEEauthorblockN{
Progga Paromita Dutta\IEEEauthorrefmark{1},
Jeba Maliha\IEEEauthorrefmark{2}, and
Md Rafiul Kabir\IEEEauthorrefmark{2}
}
\IEEEauthorblockA{
\IEEEauthorrefmark{1}Department of Computer Science, Columbia University, New York, NY, USA\\
\IEEEauthorrefmark{2}School of Engineering and Technology, Central Michigan University, Mount Pleasant, MI, USA\\
Email: pd2816@columbia.edu, \{malih1j, kabir2m\}@cmich.edu
}
}
\maketitle

\begin{abstract}
Early identification of Alzheimer’s disease (AD) remains challenging because established assessment methods can be costly, resource-intensive, or unsuitable for population-scale screening. Optical coherence tomography angiography (OCTA) provides non-invasive visualization of retinal microvasculature, but existing approaches often require diagnostic labels and provide limited measurement-level interpretation. We present an explainable OCTA pipeline that integrates annotation-aware vessel segmentation, layer-specific vascular biomarker extraction, label-free phenotyping, and measurement-grounded LLM reporting. Using 117 ROSE-1 images from 39 subjects, we apply annotation-matched segmentation models to superficial vascular complex (SVC), deep vascular complex (DVC), and combined SVC+DVC representations. The models achieve ROC-AUC values of 0.916–0.970 and Dice scores of 0.695–0.781. Six density and fractal-dimension biomarkers form subject-level profiles for exploratory clustering. Analysis of nine held-out subjects identifies an internally consistent lower-density, lower-fractal-dimension phenotype, although the absence of diagnostic labels prevents clinical interpretation. Reports generated using GPT, Gemini, and Llama are evaluated for measurement grounding, citation faithfulness, and diagnostic caution. Overall, the framework provides a transparent, non-diagnostic connection between retinal vascular measurements, exploratory phenotyping, and evidence-linked interpretation for Alzheimer’s research.

\end{abstract}
\begin{IEEEkeywords}
Alzheimer's disease, OCTA, Retinal biomarkers, LLM, Unsupervised phenotyping
\end{IEEEkeywords}

\section{Introduction}

Alzheimer's disease (AD) affects a growing population worldwide, while timely identification remains challenging. Established assessment methods, including positron emission tomography, magnetic resonance imaging, and cerebrospinal-fluid biomarkers, can be costly, resource-intensive, or procedurally burdensome for population-scale screening~\cite{ref_wagner}. The retina provides an accessible site for studying neurodegenerative vascular changes, and optical coherence tomography angiography (OCTA) enables non-invasive visualization of retinal microvasculature at capillary resolution. Previous studies have reported AD-associated reductions in vessel density and vascular branching complexity~\cite{ref_chua}.

\begin{figure}[h]
\centering
\rotatebox{-90}{%
    \includegraphics[
        height=\columnwidth,
        trim={155bp 20bp 155bp 15bp},
        clip
    ]{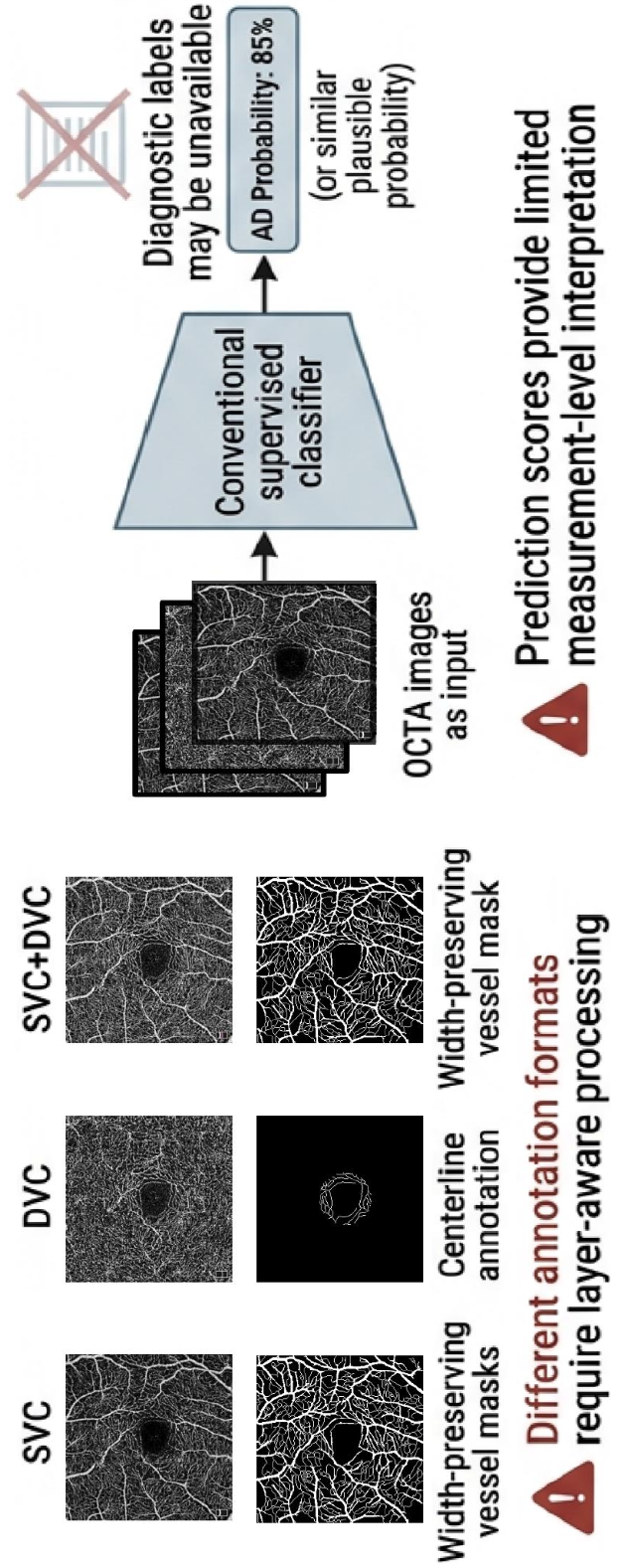}
}
\caption{Motivation for annotation-aware OCTA analysis and measurement-grounded LLM reporting}
\label{fig:motivation}
\end{figure}

Most OCTA-based AD approaches use supervised classifiers that require confirmed diagnostic labels and often produce a prediction score without identifying the vascular measurements underlying it. Their use is therefore limited in unlabeled cohorts. In addition, OCTA representations may provide different annotation formats, including width-preserving vessel masks and centerline labels, requiring layer-specific segmentation, evaluation, and biomarker definitions. Existing systems also rarely connect quantitative OCTA measurements with structured, evidence-linked subject-level interpretation. These challenges are summarized in Fig.~\ref{fig:motivation}.
We present an explainable OCTA pipeline combining annotation-aware vessel segmentation, layer-appropriate biomarker extraction, label-free phenotyping, and measurement-grounded report generation. Vessel density and fractal dimension are extracted from SVC and SVC+DVC, while normalized centerline density and fractal dimension are extracted from DVC. These measurements form a six-dimensional subject profile that supports
exploratory clustering without diagnostic labels and enables report
generation using cohort-relative $z$-scores and curated evidence. The framework is intended for transparent research analysis rather than clinical diagnosis.

The study has three objectives: (1) to develop annotation-aware segmentation and layer-appropriate biomarker extraction for heterogeneous OCTA annotations; (2) to examine whether multi-layer vascular measurements reveal internally stable phenotype structure without diagnostic labels; and (3) to generate subject-level reports grounded in measured biomarkers and curated scientific evidence. To achieve these objectives, we make three contributions. \textit{First}, it adapts the ROSE segmentation framework to heterogeneous annotation formats through two-branch models for pixel-annotated SVC and SVC+DVC images and a reduced single-branch model for centerline-annotated DVC images. \textit{Second}, it constructs a six-biomarker, multi-layer subject representation and evaluates exploratory phenotype structure through clustering and internal stability analyses. \textit{Third}, it develops and evaluates a measurement-grounded reporting pipeline across three language models, focusing on grounding accuracy, citation faithfulness, and diagnostic caution.

The remainder of this paper is organized as follows. Section~II reviews related work. Section~III presents the proposed methodology and implementation, including annotation-aware segmentation, vascular phenotyping, and LLM-based reporting. Section~IV presents the experimental results and discusses the observed errors and limitations. Section~V concludes the paper and outlines future work.
\section{Related Work}
\label{sec:related}

OCTA has been investigated as a non-invasive source of retinal vascular biomarkers for AD and mild cognitive impairment (MCI). Prior studies have reported reductions in vessel density and vascular fractal dimension, although the affected retinal plexuses vary across cohorts~\cite{ref_bulut,ref_chua}. Findings for the foveal avascular zone (FAZ) are less consistent, with studies reporting both enlargement and no significant differences~\cite{ref_bulut,ref_chua}. Such variability has been attributed to differences in populations, imaging devices, scan protocols, retinal-layer definitions, and processing methods~\cite{ref_wagner}. Consequently, no single OCTA measurement has emerged as a consistently reproducible and device-independent AD biomarker, motivating multi-layer and multi-measure vascular analysis.

Existing OCTA-based diagnostic systems generally use either handcrafted features or image-based learning. Feature-based methods combine vessel-density, FAZ, morphological, or radiomic measurements with conventional classifiers~\cite{ref_octaad1,ref_octaad2,ref_tian}. Although relatively interpretable, these methods are sensitive to segmentation and feature-selection choices and may overfit small cohorts. Image-based approaches learn directly from OCTA images or vascular maps. Xie \etal \cite{ref_xie} combined OCTA segmentation with analysis of AD- and MCI-associated microvascular changes, while Hao \etal ~\cite{ref_eyead} modeled relationships within and across vascular representations using a trustworthy-AI framework. These methods demonstrate strong discrimination but require diagnostic labels and commonly explain predictions through post-hoc attention or importance maps rather than quantified subject-level biomarkers.

Reliable biomarker analysis also depends on vessel segmentation. ROSE introduced a two-stage framework for thin- and thick-vessel segmentation with different annotation formats across vascular representations~\cite{ref_octanet}, while OCTA-500 provides a larger multi-projection dataset with multiple annotation types~\cite{ref_octa500}. However, segmentation benchmarks rarely connect annotation-aware outputs with downstream phenotyping and structured interpretation. Although retrieval-augmented generation can condition language models on external evidence~\cite{ref_rag}, retrieved evidence alone does not ensure that generated interpretations remain faithful to an individual subject's measurements. Maliha and Kabir~\cite{maliha2026llms} validated LLM-generated clinical narratives through reverse extraction, achieving 94.61\% accuracy. Their lower predictive performance than conventional machine learning supports using LLMs for explanation rather than diagnosis. 
The proposed framework addresses these gaps through annotation-aware segmentation, layer-appropriate biomarkers, label-free phenotyping, and measurement-grounded reporting without assigning an AD diagnosis.

\section{Methodology and Implementation}
\label{sec:methods}

\subsection{Dataset and Annotations}
\label{sec:dataset}

We use the publicly available ROSE-1 dataset~\cite{ref_octanet}, which contains 117 en-face OCTA images from 39 subjects. Each subject contributes three $3\times3$~mm\textsuperscript{2} fovea-centered vascular representations at a resolution of $304\times304$ pixels: the superficial vascular complex (SVC), deep vascular complex (DVC), and combined SVC+DVC representation. SVC and SVC+DVC are provided with width-preserving pixel-level vessel masks, whereas DVC is provided with centerline annotations. The public release used in this study does not include subject-level AD or healthy-control labels; therefore, all downstream analyses are exploratory and label-free.

The predefined subject-disjoint split contains 30 training subjects and nine held-out test subjects, with all three representations from each subject assigned to the same partition. Images are processed at their native resolution without resizing. During training, random rotation within $\pm10^{\circ}$ is applied jointly to each image and its corresponding annotation. The training partition is used for segmentation-model development, while the test partition is reserved for segmentation evaluation, biomarker extraction, phenotyping, and report generation. No separate validation partition is used; model checkpoints are selected after a fixed training schedule.

\subsection{OCTA Phenotyping Framework}
\label{sec:overview}
\label{sec:implementation}

\begin{figure*}[t]
\centering
\includegraphics[
  width=\textwidth,
  height=0.4\textheight,
  keepaspectratio
]{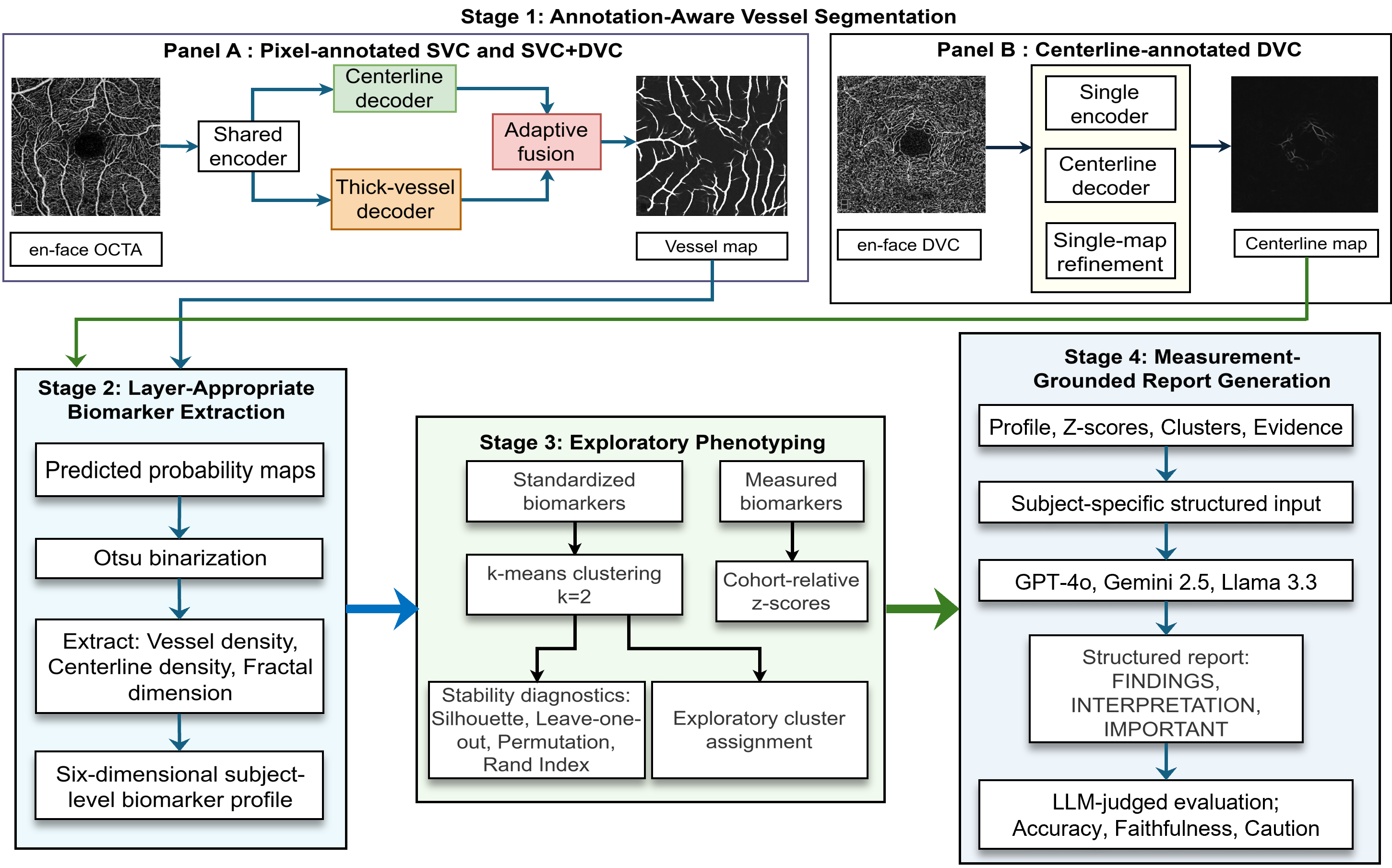}
\caption{End-to-end framework comprising annotation-aware segmentation,
layer-appropriate biomarker extraction, label-free phenotyping, and
measurement-grounded report generation.}
\label{fig:pipeline}
\end{figure*}

Fig.~\ref{fig:pipeline} summarizes the complete workflow from OCTA images to structured subject-level reports. The method consists of four stages: annotation-aware vessel segmentation, layer-appropriate biomarker extraction, exploratory unsupervised phenotyping, and measurement-grounded report generation.
For segmentation, we adapt the coarse-to-fine ROSE framework~\cite{ref_octanet} to the annotation format of each vascular representation. The pixel-annotated SVC and SVC+DVC models use a shared encoder with separate thick-vessel and centerline decoder branches. Their coarse predictions are fused during the refinement stage. Because DVC provides only centerline annotations, it is processed using a reduced single-branch encoder-decoder followed by refinement of a single centerline probability map. This design avoids imposing a thick-versus-thin target decomposition that is absent from the DVC annotations.

Predicted probability maps are binarized independently using Otsu thresholding~\cite{ref_otsu}. For SVC and SVC+DVC, vessel density (VD) is calculated as the proportion of pixels assigned to vessels. For DVC, normalized centerline density (NCD) is calculated as the proportion of pixels occupied by the predicted centerline. Fractal dimension (FD) is estimated for all three representations using box counting with box sizes of 2, 4, 8, 16, 32, and 64 pixels. 
FAZ area was also investigated using a central connected-component procedure
with morphological closing. Because qualitative quality control revealed
frequent background leakage, it was excluded from the final biomarker set;
the observed failure case is discussed in Section~\ref{sec:error}.
The retained biomarkers form the following six-dimensional representation
for subject $i$:
\begin{equation}
\begin{aligned}
\mathbf{x}_i = \big[\,
& \mathrm{VD}^{\mathrm{SVC}}_i,\; \mathrm{FD}^{\mathrm{SVC}}_i,\;
\mathrm{NCD}^{\mathrm{DVC}}_i,\; \mathrm{FD}^{\mathrm{DVC}}_i,\\
& \mathrm{VD}^{\mathrm{SVC+DVC}}_i,\; \mathrm{FD}^{\mathrm{SVC+DVC}}_i
\,\big].
\end{aligned}
\end{equation}

Each biomarker is standardized to zero mean and unit variance across the nine held-out subjects. We apply $k$-means clustering without diagnostic labels. The primary analysis uses $k = 2$ because it yielded the highest silhouette score among $k \in \{2,3,4\}$ ($0.626$ versus $0.500$ and $0.468$ for $k=3$ and $k=4$, respectively)~\cite{punhani2022binning}. Internal robustness is evaluated using Leave-one-subject-out (LOSO) re-clustering, a permutation test with 2,000 independently permuted feature sets, and cross-layer agreement \footnote{Independent feature permutation does not preserve cross-feature covariance, so this is a restricted null test rather than a significance test.}. 
Agreement between cluster partitions is measured using the adjusted Rand index~\cite{warrens2022understanding}. These analyses evaluate internal stability within the held-out cohort and do not constitute external clinical validation. 
Algorithm~\ref{alg:pipeline} summarizes the complete computational workflow of report generation.



\begin{algorithm}[t]
\footnotesize
\caption{OCTA phenotyping and reporting workflow}
\label{alg:pipeline}
\begin{algorithmic}[1]
\REQUIRE OCTA images, vessel annotations, and evidence mapping
\FOR{each vascular representation}
\STATE Train the annotation-matched model and predict test maps
\STATE Binarize maps and extract VD/NCD and FD
\ENDFOR
\STATE Standardize biomarkers; perform $k$-means and stability analysis
\STATE Compute $z$-scores and deterministic phenotype descriptions
\STATE Generate and evaluate evidence-linked reports
\end{algorithmic}
\end{algorithm}

The pipeline was implemented in Python 3.10 using PyTorch 2.5.1 with CUDA
12.1, NumPy 1.26.4, SciPy 1.15.3, and scikit-learn 1.3.2. Models were
trained on an NVIDIA RTX A6000 GPU using Adam with a learning rate of
$5\times10^{-4}$, batch size two, polynomial learning-rate decay, and 200
epochs per stage. SVC and SVC+DVC used MSE loss, whereas DVC used MSE for
coarse prediction and Dice loss for refinement. Clustering used a fixed
seed of 42.


\subsection{Evidence-Grounded LLM Report Generation} To support subject-level reporting, we implement an automated report-generation pipeline that converts quantitative OCTA-derived retinal vascular biomarkers into narrative reports. The pipeline takes patient-level biomarker values, cohort statistics, z-scores, and unsupervised cluster assignments as input, then uses three large language models to generate transparent, evidence-lined descriptions. Each report is organized into three standardized sections: Findings, Interpretation, and Important. The goal of the pipeline is not to diagnose Alzheimer’s disease, but to describe retinal vascular phenotypes in a transparent and evidence-grounded way.

\subsubsection*{Biomarker Input and Phenotype Encoding}
We begin by assembling each subject's biomarker profile from a biomarker table containing the measured value, cohort mean and standard deviation, and layer-specific $z$-score, together with the subject's unsupervised cluster assignment. This representation allows individual biomarker deviations to be interpreted alongside cluster-level phenotype information. Six biomarkers are extracted for each subject: vessel density and fractal dimension for SVC and SVC+DVC, and normalized centerline density (NCD) and fractal dimension for DVC, consistent with its centerline annotation format. Each biomarker is represented by its measured value and cohort-relative $z$-score so that the generated text remains grounded in quantitative measurements rather than qualitative impressions. Biomarkers are deterministically categorized as below the cohort mean when $z<-0.5$, above the cohort mean when $z>0.5$, and near the cohort mean otherwise. The overall profile is then assigned one of three labels: \textit{consistent-low}, \textit{consistent-high-or-near}, or \textit{mixed}. Defining these labels before report generation reduces the risk that the language model overgeneralizes the dominant pattern or overlooks biomarkers that deviate in the opposite direction.

\begin{figure}
    \centering
    \includegraphics[width=0.99\linewidth]{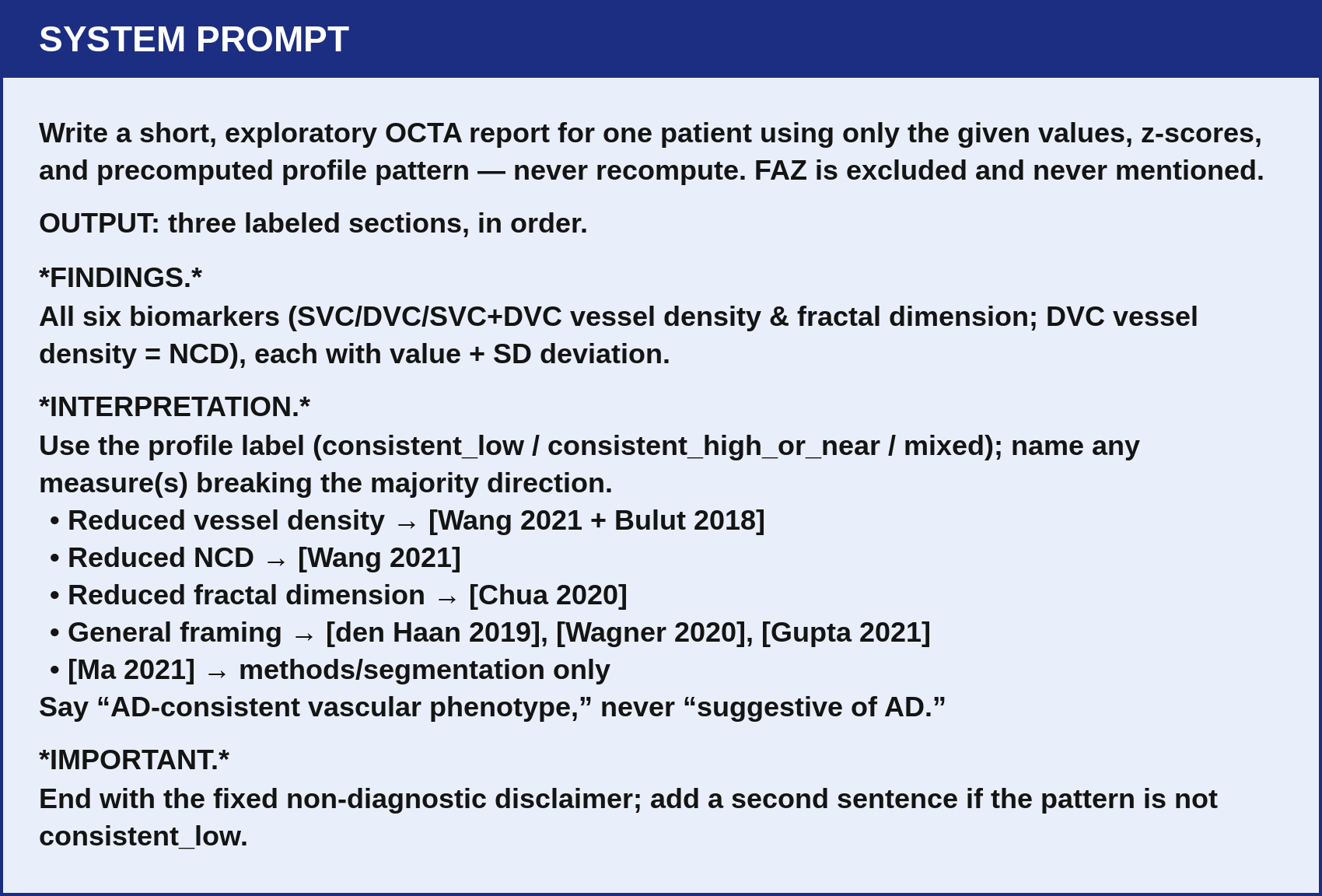}
    \caption{System Prompt for Report Generation}
    \label{fig:placeholder2}
\end{figure}

\subsubsection*{LLM Prompt for Report Generation} We use a system prompt shown in Fig~\ref{fig:placeholder2} to guide the LLM to generate reports. The prompt instructs three labeled sections: Findings, Interpretation, and Important. In Findings, the model must cover all six biomarkers, each with both its actual value and z-score-based deviations from the cohort mean. This pairing blocks unsupported or incomplete claims rather than stating only that vessel density is reduced. In interpretation, the model explains the retinal vascular pattern using the precomputed direction breakdown and phenotype label to determine how the retinal vascular pattern is explained. Consistent-low produces a lower-density/ lower-fractal phenotype description, consistent-high-or-near produces a description of that pattern's absence, and mixed requires the model to explicitly identify the profile as mixed or partial by naming the deviating biomarkers. This improves transparency and reduces the risk of misleading summaries. In Important, a fixed statement that the report is exploratory and non-diagnostic- a safeguard that matters given the unlabeled, unsupervised cohort. Diagnostic terms like ``healthy," ``normal," ``favorable," ``AD patient," or ``AD-like" are explicitly prohibited. These restrictions keep the output suited to research interpretation rather than clinical decision-making.

\subsubsection*{Literature-Grounded Citation Control}

We control the literature grounding of generated reports using a locked reference bank in which each approved source is mapped to specific biomarker patterns and permitted claims. After a report is generated, its citations are extracted and checked against the evidence mapping associated with the subject’s biomarker profile, ensuring that each claim remains traceable to an approved source. Reports are generated using three large language models: GPT, Gemini, and Llama.

\section{Experimental Results}
\label{sec:results}

\subsection{Setup and Evaluation}

We used the subject-disjoint ROSE-1 split described in Section~\ref{sec:dataset}, with 30 subjects used for segmentation-model development and nine held-out subjects reserved for evaluation. The models followed the architectures and training settings described in Section~\ref{sec:overview}. For phenotyping, $k$-means clustering was performed with 10 initializations, a fixed random seed of 42, and a primary setting of $k=2$. No diagnostic labels were used at any stage.
Segmentation performance was evaluated using pixel-wise ROC-AUC and the Dice coefficient,
\begin{equation}
\mathrm{Dice}(P,G)=\frac{2|P\cap G|}{|P|+|G|},
\end{equation}
under the layer-specific protocol described in Section~\ref{sec:implementation}. Strict pixel overlap was used for SVC and SVC+DVC, whereas DVC was evaluated using pixel-level ROC-AUC and Dice with a $3\times3$ centerline tolerance. Phenotype structure was assessed using the silhouette score, leave-one-subject-out stability, permutation testing, and cross-layer agreement measured by the adjusted Rand index. Report-generation quality was evaluated on a five-point rubric covering grounding accuracy, citation faithfulness, and diagnostic caution.

\subsection{Result Analysis of Retinal Image}

\subsubsection*{Segmentation Performance}
Table~\ref{tab:seg} summarizes the segmentation performance on the held-out ROSE-1 test set. ROC-AUC ranged from 0.916 to 0.970 across the three vascular representations. SVC achieved the
highest strict pixel-level Dice score of 0.781, while the denser combined
SVC+DVC representation achieved 0.712. The DVC model achieved a tolerance-aware centerline Dice score of 0.695 and the highest ROC-AUC of 0.970. 

\begin{table}[t]
\caption{Segmentation performance on the held-out ROSE-1 test set}
\label{tab:seg}
\centering
\footnotesize
\setlength{\tabcolsep}{4pt}
\begin{tabular*}{\columnwidth}{@{\extracolsep{\fill}}lccc@{}}
\toprule
\textbf{Representation} & \textbf{Evaluation Protocol} & \textbf{AUC} & \textbf{Dice}\\
\midrule
SVC      & Strict pixel overlap                 & 0.951 & 0.781\\
SVC + DVC  & Strict pixel overlap                 & 0.916 & 0.712\\
DVC      & AUC: pixel; Dice: $3\times3$ & 0.970 & 0.695\\
\bottomrule
\end{tabular*}
\begin{tablenotes}[flushleft]
\footnotesize
\item \textbf{Note}: Dice values are not comparable across rows. SVC and SVC+DVC use strict pixel overlap against width-preserving masks, whereas DVC uses centerline annotations with a $3\times3$ tolerance. 
\end{tablenotes}
\end{table}

\subsubsection*{Exploratory Phenotyping}

Clustering the standardized six-dimensional biomarker profiles produced two
groups containing three and six subjects. One group showed lower vessel
density in SVC and SVC+DVC, lower normalized centerline density in DVC, and
lower fractal dimension across all three representations. This group is
described as the lower-density, lower-fractal-dimension phenotype. Although the direction of
these measurements has been reported in some AD-associated OCTA studies, no
disease status can be inferred because ROSE-1 does not provide diagnostic
labels.

The silhouette score was highest for $k=2$, with values of 0.626, 0.500,
and 0.468 for $k=2$, $k=3$, and $k=4$, respectively. Leave-one-subject-out
re-clustering produced a mean ARI of 1.000. The permutation analysis yielded
an empirical $p$-value of 0.0005, indicating that the observed silhouette
was larger than those obtained after independently permuting the biomarker features. Separate clustering of the SVC, DVC, and SVC+DVC representations produced a mean cross-layer ARI of 1.000. These findings indicate strong internal consistency, but given the nine-subject cohort, they should be interpreted cautiously and validated on a larger independent dataset.

\begin{figure*}[t]
\centering
\includegraphics[
    width=1.99\columnwidth,
]{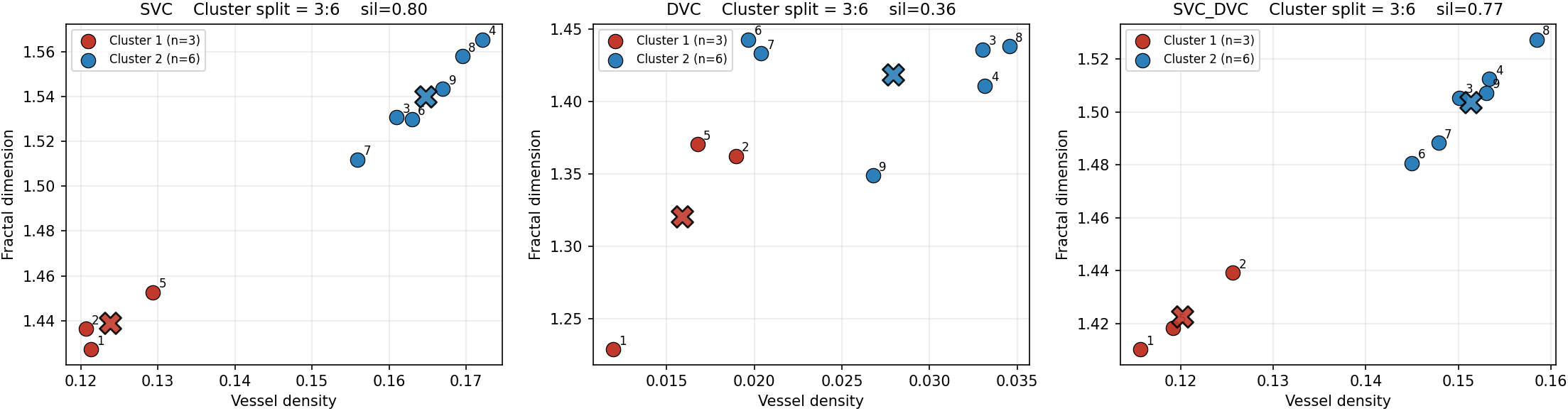}
\caption{Layer-wise phenotyping of the nine held-out subjects. The same
three subjects form the lower-density, lower-fractal-dimension group across
SVC, DVC, and SVC+DVC.}
\label{fig:clusters}
\end{figure*}

\subsection{Result Analysis of LLM-Generated Reports}

\subsubsection*{Grounding Accuracy}
We evaluate grounding accuracy to determine whether LLM-generated OCTA biomarker reports faithfully reflect the underlying subject-level data. The evaluation assesses whether each report correctly presents all six biomarkers, accurately describes their directions relative to the cohort mean and consistently interprets the overall (z)-score pattern, with particular attention to mixed profiles. An LLM-as-a-judge assigns the grounding score based on numerical accuracy and consistency with the input biomarker profile.

\subsubsection*{Citation Faithfulness}
We implement a citation faithfulness evaluation module from GPT, Gemini, and Llama/Groq for literature references accurately, checking each report against a locked reference bank and rule-based citation mapping to verify claim support, correct biomarker attribution, and the absence of invented or mismatched citations \cite{ref_bulut}. Deterministic rule checks each report for a 1–5 faithfulness score, where higher values indicate exact claim support and lower values indicate mismatched citations.

\subsubsection*{Clinical Caution}
We implement a clinical caution pipeline to determine whether LLM-generated OCTA reports preserve appropriate non-diagnostic language around Alzheimer's-related retinal vascular phenotypes, using a 1–5 caution scale. Higher scores determine whether reports preserve appropriate non-diagnostic language around Alzheimer's-related retinal vascular phenotypes, using a 1–5 caution scale. 

\subsubsection*{Report-Generation Performance}


\begin{figure}
    \centering
    \includegraphics[width=0.99\linewidth]{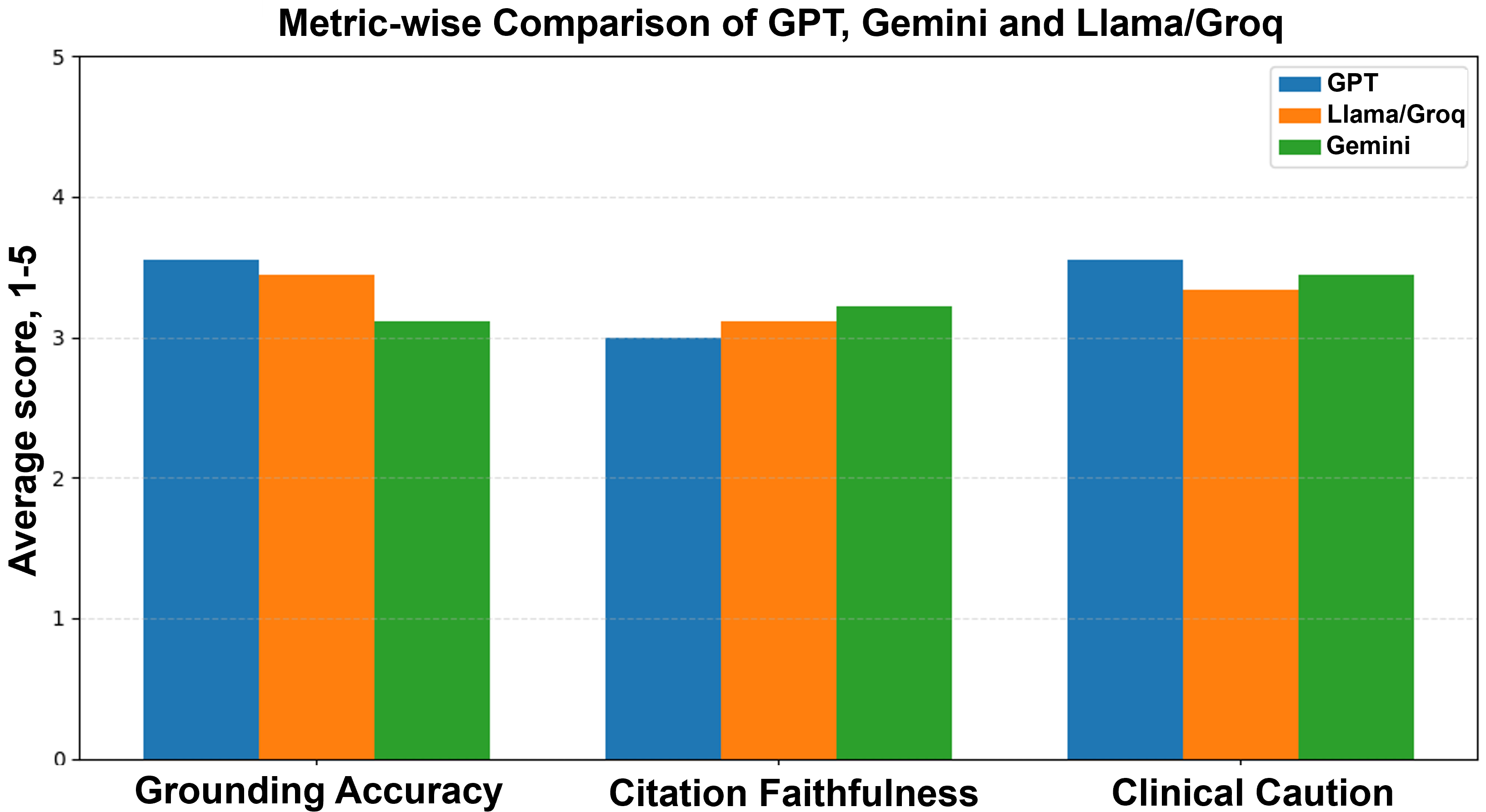}
    \caption{LLM based Evaluation Metric}
    \label{fig:EM}
\end{figure}

The three language models generated structured reports from the same subject-level biomarker profiles, cohort-relative $z$-scores, exploratory cluster assignments, and curated evidence mapping. Across GPT, Llama, and Gemini, mean grounding accuracy scores ranged from 3.11 to 3.56 on the five-point rubric. Citation-faithfulness scores ranged from 3.00 to 3.22,
while diagnostic-caution scores ranged from 3.33 to 3.56. Grounding accuracy was generally higher than citation faithfulness, suggesting that the reports reflected the measured biomarkers more reliably than they linked individual claims to specific supporting sources. Fig~\ref{fig:EM} shows the performance of the LLM evaluation metric.


\subsection{Error Analysis and Discussion}
\label{sec:error}

\begin{figure}[t]
\centering
\includegraphics[width=.85\columnwidth]{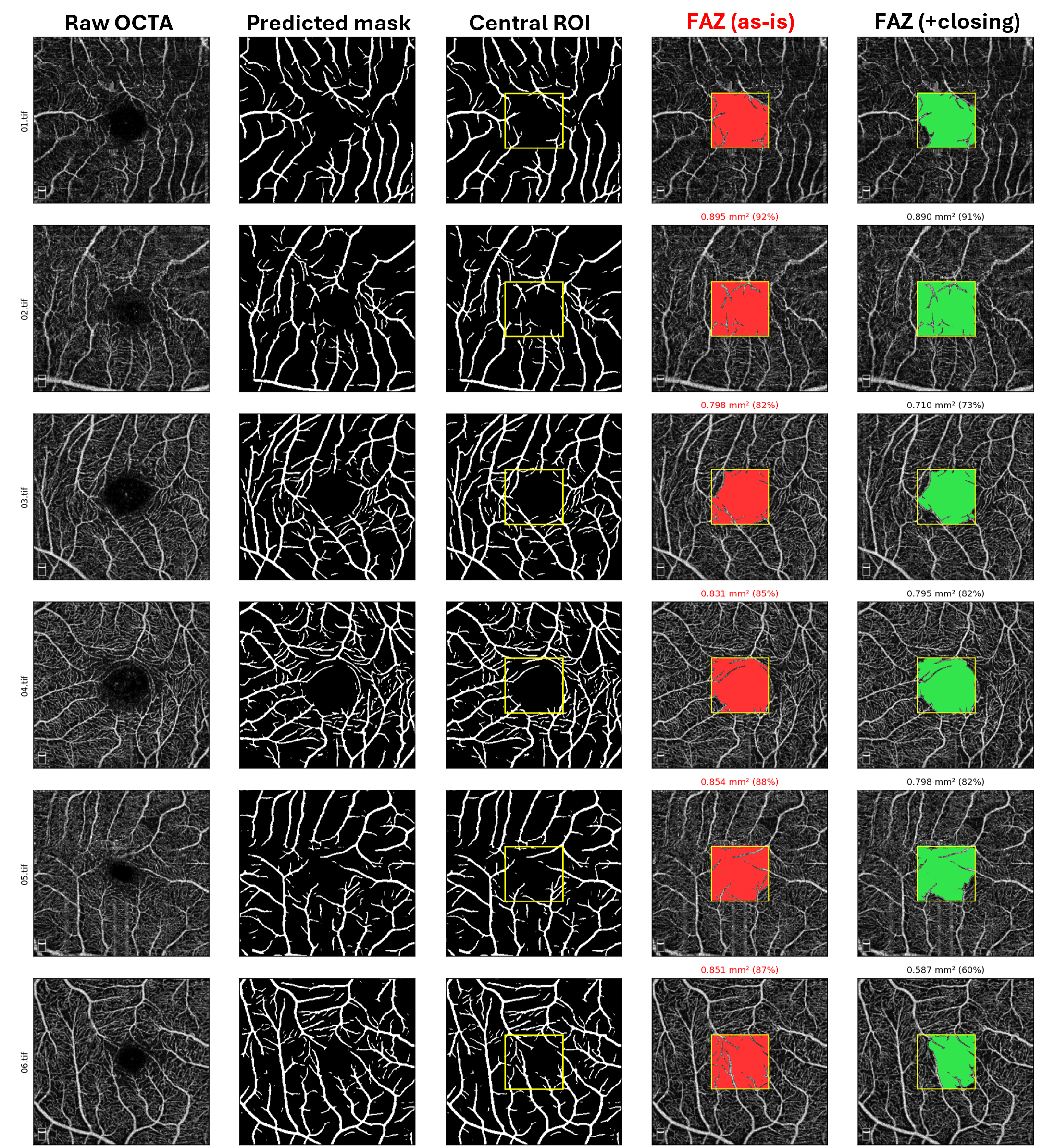}
\caption{Qualitative FAZ extraction failure caused by leakage through inter-vessel gaps, producing an implausibly large region.}
\label{fig:faz}
\end{figure}

The lower Dice score for SVC+DVC may reflect greater vessel density and overlap in the combined representation. DVC is not directly comparable with the others, as it uses centerline annotations and a tolerance-aware protocol; its higher AUC relative to Dice is expected for thin curvilinear vessels, where probability ranking stays strong while small displacements reduce exact overlap. Otsu thresholding was applied per image; although this may introduce variability, the consistent phenotype direction across all three representations suggests the separation is not layer-specific. FAZ was evaluated but excluded, as the estimated central component leaked through inter-vessel gaps, producing implausibly large regions (Fig.~\ref{fig:faz});
prior studies report inconsistent FAZ--AD associations~\cite{ref_bulut,ref_chua}.

This work targets annotation-aware integration of segmentation, biomarker extraction, and reporting, not segmentation performance alone, so we claim no superiority over alternative architectures. Controlled baselines (U-Net, Attention U-Net, nnU-Net) and ablations of the dual-branch design, refinement stage, and loss functions are natural next steps once a larger labeled cohort is available. Because the same nine subjects are used for standardization, phenotyping, and reporting, the resulting z-scores and cluster assignments are cohort-specific, and the phenotyping shows internal consistency rather than clinical validity. Feature-redundancy analysis and bootstrap uncertainty estimates would help characterize this structure, and citation faithfulness, the weakest reporting dimension, motivates claim-level citation checks and human review.

\section{Conclusion}
\label{sec:conclusion}

We presented an explainable OCTA pipeline integrating annotation-aware vessel segmentation, layer-appropriate biomarker extraction, label-free phenotyping, and measurement-grounded report generation. On the ROSE-1 dataset, the segmentation models achieved ROC-AUC values of 0.916--0.970 and Dice scores of 0.695--0.781 across SVC, DVC, and SVC+DVC. A six-dimensional biomarker representation revealed an internally consistent lower-density, lower-fractal-dimension phenotype among nine held-out subjects. However, the absence of diagnostic labels and the small evaluation cohort preclude clinical interpretation. Reports generated by three language models showed stronger measurement grounding than citation faithfulness. Overall, the framework provides a transparent, non-diagnostic link between OCTA segmentation, quantitative vascular phenotyping, and evidence-linked interpretation. 

Future work will extend the framework to labeled external cohorts, add controlled segmentation baselines, and improve FAZ extraction and citation verification. We plan to validate the report with an LLM by reverse extraction of clinical values from the generated report, and we will compare the consistency score with the source records. We will choose an LLM-generated report, as sensitive numerical patient data are not exposed in the real world.
\bibliographystyle{IEEEtran}
\bibliography{ref}

\end{document}